\pdfoutput=1

\documentclass[10pt,twocolumn,letterpaper]{article}

\usepackage[pagenumbers]{cvpr} %
\usepackage[accsupp]{axessibility}

\usepackage{graphicx}
\usepackage{amsmath}
\usepackage{amssymb}
\usepackage{booktabs}
\usepackage{multirow}
\usepackage{xcolor}
\usepackage{adjustbox}
\usepackage{bbm}

\usepackage[pagebackref,breaklinks,colorlinks]{hyperref}
\usepackage[linesnumbered,algoruled,boxed,lined]{algorithm2e}
\SetKwInOut{Parameter}{parameter}

\usepackage[capitalize]{cleveref}
\crefname{section}{Sec.}{Secs.}
\Crefname{section}{Section}{Sections}
\Crefname{table}{Table}{Tables}
\crefname{table}{Tab.}{Tabs.}

\def\eg{\emph{e.g}\onedot} 
\def\ie{\emph{i.e}\onedot}

\makeatother

\newcommand{\COMMENT}[2][.4\linewidth]{\leavevmode\hfill\makebox[#1][l]{//~#2}}
\newcommand{\tabincell}[2]{\begin{tabular}{@{}#1@{}}#2\end{tabular}}

\newcommand{\system}{ASM-Loc\xspace}

\usepackage{overpic}
\usepackage{enumitem} %
\usepackage{overpic} %
\usepackage{color}

\definecolor{turquoise}{cmyk}{0.65,0,0.1,0.3}
\definecolor{purple}{rgb}{0.65,0,0.65}
\definecolor{dark_green}{rgb}{0, 0.5, 0}
\definecolor{orange}{rgb}{0.8, 0.6, 0.2}
\definecolor{red}{rgb}{0.8, 0.2, 0.2}
\definecolor{darkred}{rgb}{0.6, 0.1, 0.05}
\definecolor{blueish}{rgb}{0.0, 0.3, .6}
\definecolor{light_gray}{rgb}{0.7, 0.7, .7}
\definecolor{pink}{rgb}{1, 0, 1}
\definecolor{greyblue}{rgb}{0.25, 0.25, 1}

\usepackage{blindtext}

\renewcommand{\paragraph}[1]{\vspace{1em}\noindent\textbf{#1}.}

\begin{document}
\title{ASM-Loc: Action-aware Segment Modeling for\\Weakly-Supervised Temporal Action Localization}

\author{Bo He$^{1}$, Xitong Yang$^{1}$, Le Kang$^{2}$, Zhiyu Cheng$^{2}$, Xin Zhou$^{2}$, Abhinav Shrivastava$^{1}$\\[0.5em]
$^{1}$University of Maryland, College Park \quad\quad $^{2}$Baidu Research, USA\\
{\tt\small \{bohe,xyang35,abhinav\}@cs.umd.edu, \{kangle01,zhiyucheng,zhouxin16\}@baidu.com}
}

\maketitle
\begin{abstract}
Weakly-supervised temporal action localization aims to recognize and localize action segments in untrimmed videos given only video-level action labels for training.
Without the boundary information of action segments, existing methods mostly rely on multiple instance learning (MIL), where the predictions of unlabeled instances (i.e., video snippets) are supervised by classifying labeled bags (i.e., untrimmed videos).
However, this formulation typically treats snippets in a video as independent instances, ignoring the underlying temporal structures within and across action segments.
To address this problem, we propose \system, a novel WTAL framework that enables explicit, action-aware segment modeling beyond standard MIL-based methods. 
Our framework entails three segment-centric components: (i) dynamic segment sampling for compensating the contribution of short actions; (ii) intra- and inter-segment attention for modeling action dynamics and capturing temporal dependencies; (iii) pseudo instance-level supervision for improving action boundary prediction.
Furthermore, a multi-step refinement strategy is proposed to progressively improve action proposals along the model training process.
Extensive experiments on THUMOS-14 and ActivityNet-v1.3 demonstrate the effectiveness of our approach, establishing new state of the art on both datasets. The code and models are publicly available at~\url{https://github.com/boheumd/ASM-Loc}. %
\end{abstract}

\vspace{-0.15in}
\section{Introduction}
\label{sec:intro}

\begin{figure}[t!]
    \centering
    \vspace{-0.2in}
    \adjincludegraphics[width=\linewidth, trim={{0.05\width} {0.05\height} {0.04\width} {0.03\height}},clip]{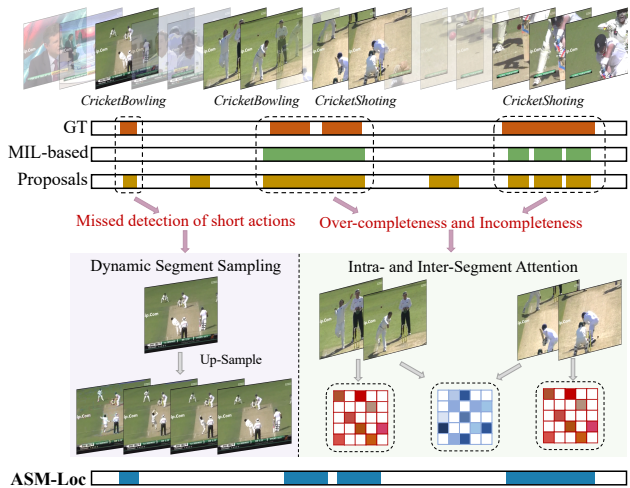}
    \caption{Action-aware segment modeling for WTAL. Our \system leverages the action proposals as well as the proposed segment-centric modules to address the common failures in existing MIL-based methods.}
    \label{fig:teaser}
    \vspace{-0.15in}
\end{figure}

Weakly-supervised temporal action localization (WTAL) has attracted increasing attention in recent years. Unlike its fully-supervised counterpart, WTAL only requires action category annotation at the video level, which is much easier to collect and more scalable for building large-scale datasets.                                
To tackle this problem, recent works~\cite{wang2017untrimmednets,nguyen2018weakly,paul2018w,liu2019completeness,luo2020weakly,lee2020background,zhai2020two,islam2021hybrid,zhang2021cola,yang2021uncertainty,liu2021blessings,huang2021foreground} mostly rely on the multiple instance learning (MIL) framework~\cite{maron1998framework}, where the entire untrimmed video is treated as a labeled bag containing multiple unlabeled instances (i.e., video frames or snippets). The action classification scores of individual snippets are first generated to form the temporal class activation sequences (CAS) and then aggregated by a top-$k$ mean mechanism to obtain the final video-level prediction~\cite{paul2018w,islam2020weakly,lee2020background,islam2021hybrid}. 

While significant improvement has been made in prior work, there is still a huge performance gap between the weakly-supervised and fully-supervised settings. 
One major challenge is localization completeness, where the models tend to generate incomplete or over-complete action segments due to the inaccurate predictions of action boundaries.
Another challenge is the missed detection of short action segments, where the models are biased towards segments with longer duration and produce low-confidence predictions on short actions.
Figure~\ref{fig:teaser} demonstrates an example of these two common errors.
Although these challenges are inherently difficult due to the lack of segment-level annotation, we argue that the absence of \textit{segment-based modeling} in existing MIL-based methods is a key reason for the inferior results.
In particular, these MIL-based methods treat snippets in a video as \textit{independent} instances, where their underlying temporal structures are neglected in either the feature modeling or prediction stage.

In this paper, we propose a novel framework that enables explicit, \textbf{a}ction-aware \textbf{s}egment \textbf{m}odeling for weakly-supervised temporal action \textbf{loc}alization, which we term \system.
To bootstrap segment modeling, we first generate action proposals using the standard MIL-based methods. These proposals provide an initial estimation of the action locations in the untrimmed video as well as their duration.
Based on the action proposals, we introduce three segment-centric modules that correspond to the three stages of a WTAL pipeline, i.e., the feature extraction stage, the feature modeling stage and the prediction stage.

First, a \textbf{dynamic segment sampling} module is proposed to balance the contribution of short-range and long-range action segments. As shown in Figure~\ref{fig:teaser}, action proposals with short duration are up-sampled along the temporal dimension, with the scale-up ratios dynamically computed according to the length of the proposals.
Second, \textbf{intra- and inter-segment attention} modules are presented to capture the temporal structures within and across action segments at the feature modeling stage. Specifically, the intra-segment attention module utilizes self-attention \textit{within} action proposals to model action dynamics and better discriminate foreground and background snippets. On the other hand, the inter-segment attention module utilizes self-attention \textit{across} different actions proposals to capture the relationships, facilitating the localization of action segments that involve temporal dependencies (e.g., ``\textit{CricketBowling}" is followed by ``\textit{CricketShotting}" in Figure~\ref{fig:teaser}).
Note that both attention modules are segment-centric, which is critical to suppress the negative impact of noisy background snippets in untrimmed videos.
Third, a \textbf{pseudo instance-level loss} is introduced to refine the localization result by providing fine-grained supervision. The pseudo instance-level labels are derived from the action proposals, coupled with uncertainty estimation scores that mitigate the label noise effects. 
Finally, a multi-step proposal refinement is adopted to progressively improve the quality of action proposals, which in turn boosts the localization performance of our final model.

We summarize our main contributions as follows:
\begin{itemize}
    \vspace{-0.5em}
    \item We show that segment-based modeling can be utilized to narrow the performance gap between the weakly-supervised and supervised settings, which has been neglected in prior MIL-based WTAL methods.
    \vspace{-0.5em}
    \item We introduce three novel segment-centric modules that enable action-aware segment modeling in different stages of a WTAL pipeline.
    \vspace{-0.5em}
    \item We provide extensive experiments to demonstrate the effectiveness of each component of our design. Our \system establishes new state of the art on both THUMOS-14 and ActivityNet-v1.3 datasets.
\end{itemize}

\section{Related works}
\label{sec:related}

\paragraph{Temporal Action Localization (TAL)}
Compared with action recognition~\cite{carreira2017quo,feichtenhofer2019slowfast,Yang_2021_CVPR,he2020gta,yang2021hierarchical,wu2020dynamic,wang2021bevt}, TAL is an more challenging task for video understanding.
Current fully-supervised TAL methods can be categorized into two groups:
the anchor-based methods~\cite{chao2018rethinking,shou2016temporal,xu2017r,zhao2017temporal} perform boundary regression based on pre-defined action proposals, while the anchor-free methods~\cite{lin2018bsn,lin2019bmn,lin2020fast} directly predict boundary probability or actionness scores for each snippet in the video, and then employ a bottom-up grouping strategy to match pairs of start and end for each action segment.
All these methods require precise temporal annotation of each action instance, which is labor-intensive and time-consuming.

\paragraph{Weakly-supervised Temporal Action Localization}
Recently, the weakly supervised setting, where only video-level category labels are required during training, has drawn increasing attention from the community~\cite{wang2017untrimmednets,singh2017hide,nguyen2018weakly,shou2018autoloc,paul2018w,liu2019completeness,liu2019weakly,nguyen2019weakly,luo2020weakly,lee2020background,zhai2020two,shi2020weakly,alwassel2019refineloc,luo2021action,huang2021foreground,islam2021hybrid,zhang2021cola,yang2021uncertainty,liu2021blessings}.
Specifically, UntrimmedNet~\cite{wang2017untrimmednets} is the first to introduce the multiple instance learning (MIL) framework to tackle this problem, which selects foreground snippets and groups them as action segments.
STPN~\cite{nguyen2018weakly} improves UntrimmedNet by adding a sparsity loss to enforce the sparsity of selected snippets.
CoLA~\cite{zhang2021cola} utilizes contrastive learning to distinguish the foreground and background snippets.
UGCT~\cite{yang2021uncertainty} proposes an online pseudo label generation with uncertainty-aware learning mechanism to impose the pseudo label supervision on the attention weight.
All these MIL-based methods treat each snippet in the video individually, neglecting the rich temporal information at the segment-level.
In contrast, our \system focuses on modeling segment-level temporal structures for WTAL, which is rarely explored in prior work.

\paragraph{Pseudo Label Guided Training}
Using pseudo labels to guide model training has been widely adopted in vision tasks with weak or limited supervision. 
In weakly supervised object detection, one of the seminal directions is self-training~\cite{zou2019confidence,zhang2018w2f,li2016weakly,ren2020instance}, which first trains a teacher model and then the predictions with high confidence are used as instance-level pseudo labels to train a final detector.
Similarly, in semi-supervised learning~\cite{lee2013pseudo,laine2016temporal,berthelot2019mixmatch,sohn2020fixmatch,weng2021semi} and domain adaptation~\cite{saito2017asymmetric,liang2019exploring,das2018graph}, models are first trained on the labeled / source dataset and then used to generate pseudo labels for the unlabeled / target dataset to guide the training process.

Similar to these works, our \system utilizes pseudo segment-level labels (i.e., action proposals) to guide our training process in the WTAL task. 
However, we do not limit our approach to using pseudo labels for supervision only.
Instead, we leverage the action proposals in multiple segment-centric modules, such as dynamic segment sampling, intra- and inter-segment attention.

\section{WTAL Base Model}
\label{sec:baseline}

\begin{figure*}[t!]
    \centering
    \vspace{-0.25in}
    \includegraphics[width=0.97\textwidth]{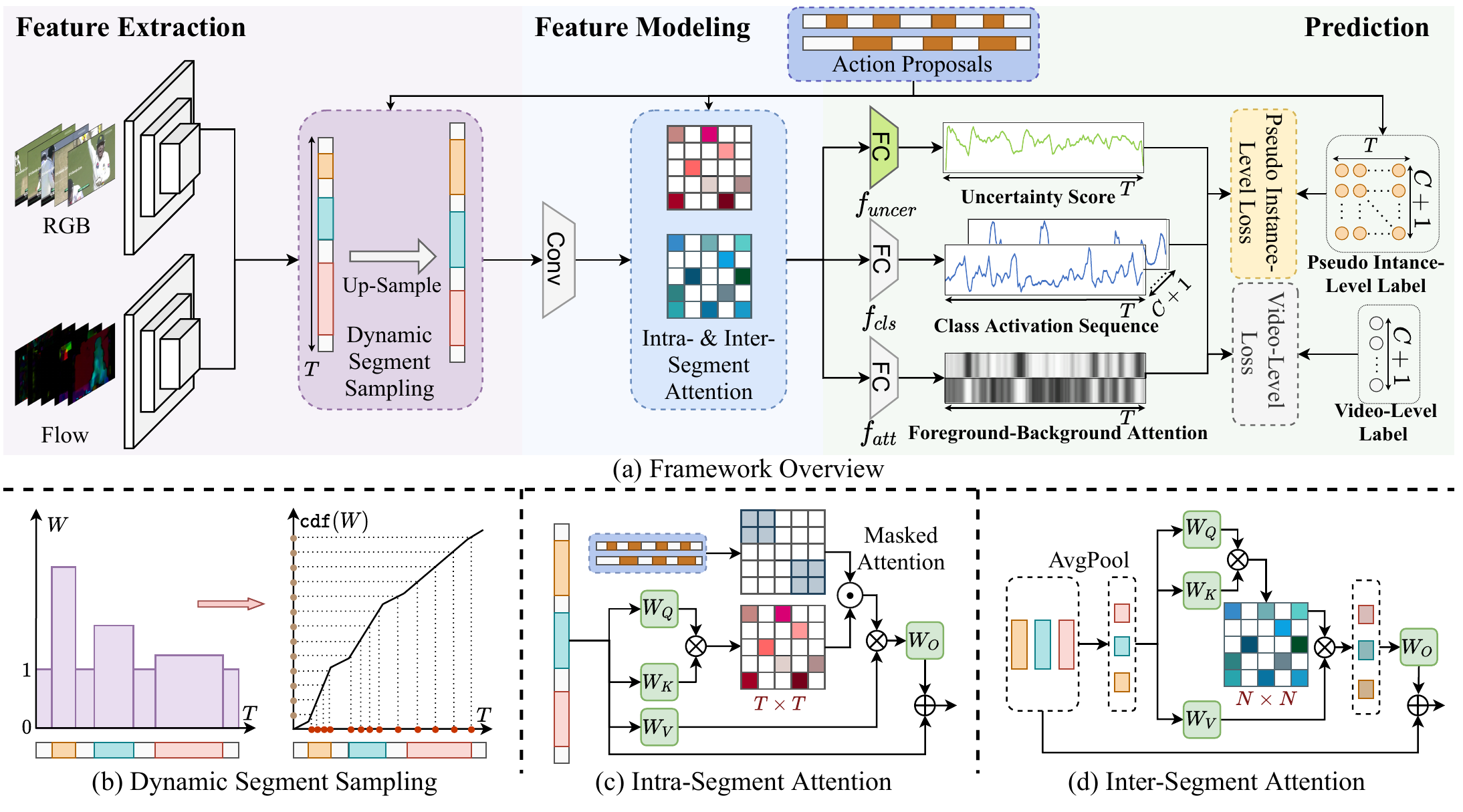}
    \vspace{-0.05in}
    \caption{(a) Framework Overview. The \textcolor{gray}{gray} modules indicate the components of the base model (\eg conv and FC),  while the others are our action-aware segment modeling modules. (b) Dynamic segment sampling is based on the cumulative distribution of the sampling weight vector $W$. The red dots on the $T$-axis represent the final sampled timesteps. Shorter action segments have higher scale-up ratios. (c) Intra-segment attention applies self-attention within each action proposal. (d) Inter-segment attention applies self-attention among all proposals in a video. $\bigodot$, $\bigotimes$ and $\bigoplus$ denote element-wise multiplication, matrix multiplication, and element-wise addition. $T$, $N$ are the number of snippets and action proposals, respectively.}
    \label{fig:model}
    \vspace{-0.15in}
\end{figure*}

WTAL aims to recognize and localize action segments in untrimmed videos given only video-level action labels during training.
Formally, let us denote an untrimmed training video as $V$ and its ground-truth label as $y\in \mathbb{R}^{C}$, where $C$ is the number of action categories. Note that $y$ could be a multi-hot vector if more than one action is present in the video and is normalized with the $l_1$ normalization. The goal of temporal action localization is to generate a set of action segments $\mathcal{S} = \{(s_i, e_i, c_i, q_i)\}_{i=1}^{I}$ for a testing video, where $s_i, e_i$ are the start and end time of the $i$-th segment and $c_i, q_i$ are the corresponding class prediction and confidence score.

Most existing WTAL methods~\cite{wang2017untrimmednets,nguyen2018weakly,paul2018w,liu2019completeness,luo2020weakly,lee2020background,zhai2020two,islam2021hybrid,zhang2021cola,yang2021uncertainty,liu2021blessings,huang2021foreground} employ the multiple instance learning (MIL) formulation. A typical pipeline of MIL-based methods consists of three main stages (depicted in Figure~\ref{fig:model}): (i) The \textit{feature extraction} stage takes the untrimmed RGB videos and optical flow as input to extract snippet-level features using pre-trained backbone networks. (ii) The \textit{feature modeling} stage transforms the extracted features to the task-oriented features by performing temporal modeling. (iii) The \textit{prediction} stage generates class probabilities and attention weights for each time step and computes video-level loss following the MIL formulation during training.
In the following subsections, we review the common practices of these three stages and present our base model in detail.

\subsection{Feature Extraction and Modeling}
\label{sec:feature}
Following the recent WTAL methods~\cite{nguyen2018weakly,liu2019completeness,nguyen2019weakly,alwassel2019refineloc,yang2021uncertainty}, we first divide each untrimmed video into non-overlapping 16-frame snippets, and then apply a Kinetics-400 pre-trained I3D model~\cite{carreira2017quo} to extract features for both RGB and optical flow input.
After that, the RGB and optical flow features are concatenated along the channel dimension to form the snippet-level representations $F\in \mathbb{R}^{T\times D}$, where $T$ is the number of snippets in the video and $D=2048$ is the feature dimensionality. Following ~\cite{liu2019completeness,lee2020background,zhang2021cola,qu2021acm}, the features are then fed into a  temporal convolution layer and the ReLU activation for feature modeling: $X = \text{ReLU}(\text{conv}(F))$.

\subsection{Action Prediction and Training Losses}
\label{sec:loss}
Given the embedded features $X$, a fully-connected (FC) layer is applied to predict the temporal class activation sequence (CAS) $P \in \mathbb{R}^{T\times (C+1)}$, where $C+1$ denotes the number of action categories plus the background class.
To better differentiate the foreground and background snippets, a common strategy~\cite{nguyen2018weakly,liu2019completeness,zhai2020two} is to introduce an additional attention module that outputs the attention weights for each time step of the untrimmed video.
Following~\cite{alwassel2019refineloc,qu2021acm}, we generate the attention weights $A\in \mathbb{R}^{T\times 2}$ using an FC layer, where the two weight values at each time step are normalized by the softmax operation to obtain the foreground and background attention weights, respectively.
Finally, the CAS and the attention weights are combined to get the attention weighted CAS: $\hat{P}^{m}(c) = P(c) \odot A^{m}, m\in \{\text{fg},\text{bg}\}$, where $c$ indicates the class index and $\odot$ denotes element-wise multiplication.

Following the MIL formulation, the video-level classification score is generated by the top-$k$ mean strategy~\cite{paul2018w,lee2020background,islam2021hybrid}.
For each class $c$, we take the $k$ largest values of the attention weighted CAS and compute their averaged value: $\hat{p}^{m}(c) = \frac{1}{k}\sum \text{Top-k}(\hat{P}^{m}(c))$. Softmax normalization is then performed across all classes to obtain the attention weighted video-level action probabilities. We adopt three video-level losses in such a weakly-supervised setting.

\paragraph{Foreground loss} 
To guide the training of video-level action classification, we apply the cross-entropy loss between the foreground-attention weighted action probabilities $\hat{p}^\text{fg}$ and the video-level action label $y^\text{fg} = [y; 0]$, written as:
\vspace{-0.08in}
\begin{equation}
    \mathcal{L}^\text{fg}=-\sum_{c=1}^{C+1}y^\text{fg}(c)\log\hat{p}^\text{fg}(c).
    \vspace{-0.15in}
\end{equation}

\paragraph{Background loss} 
 To ensure that the negative instances in the untrimmed video are predicted as the background class, we regularize the background-attention weighted action probabilities $\hat{p}^\text{bg}$ with an additional background loss~\cite{nguyen2019weakly,qu2021acm}. Specifically, we compute the cross-entropy between $\hat{p}^\text{bg}$ and the background class label $y^\text{bg}$: 
\vspace{-0.03in}
\begin{equation}
    \mathcal{L}^\text{bg}=-\sum_{c=1}^{C+1}y^\text{bg}(c)\log\hat{p}^\text{bg}(c),
    \vspace{-0.03in}
\end{equation}
where $y^\text{bg}(C+1)=1$ and $y^\text{bg}(c)=0$ for all other $c$.

\paragraph{Action-aware background loss}
Although no action is taking place in background snippets, we argue that rich context information is still available to reflect the actual action category label.
As an example in Figure~\ref{fig:qualitative}(c), even though the background frames are stationary with only a billiard table, one can still expect the existence of the action category ``\textit{Billiard}" somewhere in the video. Therefore, the background instances are related to not only the background class label but also the action class label. 

Based on this observation, we formulate the action-aware background loss as the cross-entropy loss between the background-attention weighted action probabilities $\hat{p}^\text{bg}$ and the video-level action label $y^\text{fg}$:
\vspace{-0.03in}
\begin{equation}
    \mathcal{L}^\text{abg}=-\sum_{c=1}^{C+1}y^\text{fg}(c)\log\hat{p}^\text{bg}(c).
    \vspace{-0.03in}
\end{equation}

The total video-level loss for our base model is the weighted combination of all three losses:
\begin{equation}
    \mathcal{L}^\text{vid}=\lambda_\text{fg} \mathcal{L}^\text{fg} + \lambda_\text{bg} \mathcal{L}^\text{bg} + \lambda_\text{abg} \mathcal{L}^\text{abg},
\end{equation}
where $\lambda_\text{fg}, \lambda_\text{bg}$ and $\lambda_\text{abg}$ are trade-off parameters for balancing the contribution of the three losses.

\subsection{Discussion}
As discussed in Sec.~\ref{sec:intro}, our base model follows the MIL formulation and neglects the temporal structures among video snippets.
Nevertheless, the prediction results generated by the base model still provide a decent estimation of the action locations and durations in the untrimmed video, which can serve as a bootstrap for our segment modeling process.
In particular, we generate the initial action proposals based on the prediction results of the base model: $\mathcal{S} \mapsto \tilde{\mathcal{S}} = \{(s_n, e_n, c_n)\}_{n=1}^N$, where $s_n$, $e_n$ and $c_n$ denote the start time, the end time, and the predicted category label of the $n$-th action proposal, respectively. More details on generating action proposals are available in the supplementary material.
The main focus of our work is to leverage the action proposals for segment-level temporal modeling, as described in the following section.

\section{Action-aware Segment Modeling}
\label{sec:segment_modeling}
Figure~\ref{fig:model}(a) illustrates an overview of our \system framework. Given the action proposals generated by the base model, we introduce action-aware segment modeling into all three stages of the WTAL pipeline: dynamic segment sampling in the feature extraction stage (Sec.~\ref{sec:dss}), intra- and inter-segment attention in the feature modeling stage (Sec.~\ref{sec:att}) and pseudo instance-level supervision in the prediction stage (Sec.~\ref{sec:pseudo}).
A multi-step proposal refinement is adopted to progressively improve the action proposals and the localization results, as discussed in Sec.~\ref{sec:iterative}.

\subsection{Dynamic Segment Sampling}
\label{sec:dss}
Action segments in an untrimmed video may have various duration, ranging from less than 2 seconds to more than 1 minute. Intuitively, short actions have small temporal scales, and therefore, their information is prone to loss or distortion throughout the feature modeling stage. As shown in Table~\ref{tab:dynamic_sampling}, we observe that models are indeed biased towards the segments with longer duration and produce lower confidence scores on short segments, resulting in missed detection or inferior localization results. 
Similar observations are in object detection, where smaller objects have worse detection performance than larger ones~\cite{lin2017feature,singh2018sniper}.

In order to address this problem in the WTAL setting, we propose a novel segment sampling module that dynamically up-samples action proposals according to their estimated duration.
Formally, we first initialize a sampling weight vector $W \in \mathbb{R}^{T}$ with values equal to 1 at all time steps.
Then, we compute the updated sampling weight for short proposals with duration less than a pre-defined threshold $\gamma$:
\begin{equation}
    W[s_n:e_n] = \dfrac{\gamma}{e_n - s_n}, \ \ \text{ if } (e_n - s_n) \leq \gamma,
\end{equation}
where $s_n, e_n$ denote the start and end time of the $n$-th action proposal.
The sampling procedure is based on the Inverse Transform Sampling method as shown in Figure~\ref{fig:model}(b). The intuition is to sample snippets with frame rates proportional to their sampling weights $W$.
We first compute the cumulative distribution function (CDF) of the sampling weights $f_W=\texttt{cdf}(W)$, then uniformly sample $T$ timesteps from the inverse of the CDF:  $\{x_i = f_W^{-1}(i)\}_{i=1}^T$. 
In this way, the scale-up ratio of each proposal is dynamically computed according to its estimated duration. 
We apply linear interpolation when up-sampling is needed.

\subsection{Intra- and Inter-Segment Attention}
\label{sec:att}
\vspace{-0.15in}
\paragraph{Intra-Segment Attention}
Action modeling is of central importance for accurate action classification and temporal boundary prediction.
Recent work~\cite{bertasius2021space,he2020gta} applies temporal attention globally on trimmed videos for action recognition and achieves impressive performance. 
However, untrimmed videos are usually dominated by irrelevant background snippets which introduce extra noise to the action segment modeling process.
Motivated by this observation, we propose the intra-segment attention module that performs self-attention \textit{within} each action proposal. 

We formulate this module using a masked attention mechanism, as shown in Figure~\ref{fig:model}(c).
Specifically, an attention mask $M \in \mathbb{R}^{T\times T}$ is defined to indicate the foreground snippets corresponding to different action proposals. The attention mask is first initialized with $0$ at all entries and assigned $M[s_n:e_n, s_n:e_n]=1$ for all proposals.
The attention mask is then applied to the attention matrix computed by the standard self-attention approach~\cite{vaswani2017attention,wang2018non}:
\begin{align}
    Q&=XW_Q,\  K=XW_K,\  V = XW_V, \label{eq:attention1} \\
    A_{i,j} &= \dfrac{M_{i,j} \text{exp}(Q_iK_j^T/\sqrt{D})}{{\sum_k M_{i,k} \text{exp}(Q_iK_k^T/\sqrt{D})}} \label{eq:attention2}\\
    Z&=X+\texttt{BN}(AV W_O), \label{eq:attention3}
\end{align}
where $W_Q,W_K,W_V,W_O\in \mathbb{R}^{D\times D}$ are the linear projection matrices for generating the query, key, value and the output. Multi-head attention~\cite{vaswani2017attention} is also adopted to improve the capacity of the attention module.
In this way, we explicitly model the temporal structures within each action proposal, avoiding the negative impact of the irrelevant and noisy background snippets.

\smallskip
\noindent\textbf{Inter-Segment Attention.} Action segments in an untrimmed video usually involve temporal dependencies with each other.
For example, ``\textit{CricketBowling}" tends to be followed by ``\textit{CricketShotting}", while ``\textit{VolleyballSpiking}" usually repeats multiple times in a video.
Capturing these dependencies and interactions among action segments can therefore improve the recognition and localization performance.

Similar to the intra-segment attention module, we leverage a self-attention mechanism to model the relationships across multiple action proposals.
As shown in Figure~\ref{fig:model}(d), we first aggregate the snippet-level features within each action proposal by average pooling on the temporal dimension $\hat{X}_n = \frac{1}{e_n-s_n+1}\sum_{t=s_n}^{e_n}X(t)$.
The multi-head self-attention is then applied on all segment-level features $\{\hat{X}_n\}_{n=1}^N$ to model the interactions between different action proposal pairs. The output features are replicated along the time axis and added to the original feature $X$ in a residual manner.

\subsection{Pseudo Instance-level Loss}
\label{sec:pseudo}
Due to the absence of segment-level annotation, standard MIL-based methods only rely on video-level supervision provided by the video-level action category label.
To further refine the localization of action boundaries, we leverage the pseudo instance-level label provided by the action proposals and propose a pseudo instance-level loss that offers more fine-grained supervision than the video-level losses.

Given the action proposals $\tilde{\mathcal{S}}=\{s_n,e_n,c_n\}_{n=1}^{N}$, we construct the pseudo instance-level label  $\tilde{Q}\in \mathbb{R}^{T\times (C+1)}$ by assigning action labels to the snippets that belong to the action proposals and assigning the background class label to all other snippets:
\begin{equation}
    \tilde{Q}_{t}(c)=\left\{
    \begin{array}{ll}
        1, \text{ if } \exists n,  t \in [s_n,e_n] \text{ and } c=c_n\\
        1, \text{ if } \forall n,  t \notin [s_n,e_n] \text{ and } c=C+1 \\
        0, \text{ otherwise }
    \end{array}
    \right. \\
\end{equation}
Note that $\tilde{Q}$ is also normalized with the $l_1$ normalization.

As the action proposals are generated from the model prediction, it is inevitable to produce inaccurate pseudo instance-level labels. To handle the label noise effects, we follow the recent work~\cite{kendall2017uncertainties,yang2021uncertainty,song2020learning,zheng2021rectifying} and introduce an uncertainty prediction module that guides the model to learn from noisy pseudo labels.
Specifically, we employ an FC layer to output the uncertainty score $U\in \mathbb{R}^{T}$, which is then used to re-weight the pseudo instance-level loss at each time step. Intuitively, instances with high uncertainty scores are limited from contributing too much to the loss. Coupled with uncertainty scores, the pseudo instance-level loss can be written as  the averaged cross-entropy between the temporal CAS $P$ and the pseudo instance-level label $\tilde{Q}$:
\begin{equation}
    \mathcal{L}_\text{ins} = \dfrac{1}{T}\sum_{t=1}^{T} \exp(-U_t)\left(-\sum_{c=1}^{C+1}\tilde{Q}_t(c)\log(P_t(c))\right) + \beta U_t
\end{equation}
where $\beta$ is a hyper-parameter for the weight decay term, which prevents the uncertainty prediction module from predicting infinite uncertainty for all time steps (and therefore zero loss). 

\subsection{Multi-step Proposal Refinement}
\label{sec:iterative}
\vspace{-0.05in}
Action proposals play an important role in action-aware modeling. As discussed in Sec.~\ref{sec:ablation}, the quality of proposals is positively correlated with the performance of multiple components in our approach.
While our initial action proposals are obtained from the base model, it is intuitive to leverage the superior prediction results generated by our \system to generate more accurate action proposals.
Based on this motivation, we propose a multi-step training process that progressively refines the action proposals via multiple steps.

As a bootstrap of segment modeling, we first train the base model (Sec.~\ref{sec:baseline}) for $E$ epochs and obtain the initial action proposals $\tilde{\mathcal{S}}_0$.
After that, we train our \system for another $E$ epochs and obtain the refined action proposals $\tilde{\mathcal{S}}_1$ with a more accurate estimation of the action location and duration.
The same process can be applied for multiple steps until the quality of action proposals is converged. 
The complete multi-step proposal refinement process is summarized in Alg.~\ref{alg:refine}.
Finally, we train our \system using the refined proposals $\tilde{\mathcal{S}}$ until the model is converged.

\begin{table*}[t]
\centering
\vspace{-0.3in}
\caption{Comparison with state-of-the-art methods on THUMOS-14 dataset. The average mAPs are computed under the IoU thresholds [0.1,0.1,0.7]. UNT and I3D are abbreviations for UntrimmedNet features and I3D features, respectively.
}
\vspace{-0.1in}
\resizebox{.82\textwidth}{!}{
\renewcommand{\arraystretch}{1.1}
    \begin{tabular*}{\linewidth}{@{\extracolsep{\fill}\;}cllcccccccc}
    \toprule
    \multirow{2}{*}{\textbf{Supervision}}       & \multicolumn{1}{l}{\multirow{2}{*}{\textbf{Method}}} &  \multicolumn{1}{c}{\multirow{2}{*}{\textbf{Publication}}} &
    \multicolumn{8}{c}{\textbf{mAP@IoU (\%)}} \\
    \cmidrule{4-11}
    & & \multicolumn{1}{c}{}  & 0.1  & 0.2  & 0.3  & 0.4  & 0.5  & 0.6  & 0.7 & AVG   \\
    \midrule
    \multirow{5}*{\tabincell{c}{Full\\(-)}} 
    & SSN~\cite{zhao2017temporal}  & \emph{ICCV 2017} & 66.0 & 59.4 & 51.9 & 41.0 & 29.8 & - & - & - \\
    & TAL-Net~\cite{chao2018rethinking} & \emph{CVPR 2018} & 59.8 & 57.1 & 53.2 & 48.5 & 42.8 & 33.8 & 20.8 & 45.1\\
    & GTAN~\cite{long2019gaussian} & \emph{CVPR 2019} & 69.1 & 63.7 & 57.8 & 47.2 & 38.8 & - & - & - \\
    & P-GCN~\cite{zeng2019graph} & \emph{ICCV 2019} & 69.5 & 67.8 & 63.6 & 57.8 & 49.1 & - & - & - \\
    & VSGN~\cite{zhao2021video} & \emph{ICCV 2021} & - & - & 66.7 & 60.4 & 52.4 & 41.0 & 30.4 & - \\
    \midrule
    \midrule
    \multirow{3}*{\tabincell{c}{Weak\\(UNT)}} & AutoLoc~\cite{shou2018autoloc} & \emph{ECCV 2018} & - & - & 35.8 & 29.0 & 21.2 & 13.4 & 5.8 & - \\
    & CleanNet~\cite{liu2019weakly} & \emph{ICCV 2019} & - & - & 37.0 & 30.9 & 23.9 & 13.9 & 7.1 & - \\
    & Bas-Net~\cite{lee2020background} & \emph{AAAI 2020} & - & - & 42.8 & 34.7 & 25.1 & 17.1 & 9.3 & - \\
    \midrule
    \multirow{9}*{\tabincell{c}{Weak\\(I3D)}} & STPN~\cite{nguyen2018weakly} & \emph{CVPR 2018} & 52.0 & 44.7 & 35.5 & 25.8 & 16.9 & 9.9 & 4.3 & 27.0\\
    & CMCS~\cite{liu2019completeness}  & \emph{CVPR 2019} & 57.4 & 50.8 & 41.2 & 32.1 & 23.1 & 15.0 & 7.0 & 32.4\\
    & WSAL-BM~\cite{nguyen2019weakly} & \emph{ICCV 2019} & 60.4 & 56.0 & 46.6 & 37.5 & 26.8 & 17.6 & 9.0 & 36.3\\
    & DGAM~\cite{shi2020weakly} & \emph{CVPR 2020} & 60.0 & 54.2 & 46.8 & 38.2 & 28.8 & 19.8 & 11.4 & 37.0\\
    & TSCN~\cite{zhai2020two} & \emph{ECCV 2020} & 63.4 & 57.6 & 47.8 & 37.7 & 28.7 & 19.4 & 10.2 & 37.8\\
    & ACM-Net~\cite{qu2021acm} & \emph{TIP 2021} & 68.9 & 62.7 & 55.0 & 44.6 & 34.6 & 21.8 & 10.8 & 42.6\\
    & CoLA~\cite{zhang2021cola} & \emph{CVPR 2021} & 66.2 & 59.5 & 51.5 & 41.9 & 32.2 & 22.0 & 13.1 & 40.9\\
    & UGCT~\cite{yang2021uncertainty} & \emph{CVPR 2021} & 69.2 & 62.9 & 55.5 & 46.5 & 35.9 & 23.8 & 11.4 & 43.6\\
    & AUMN~\cite{luo2021action} & \emph{CVPR 2021} & 66.2 & 61.9 & 54.9 & 44.4 & 33.3 & 20.5 & 9.0 & 41.5\\
    & FAC-Net~\cite{huang2021foreground} & \emph{ICCV 2021} & 67.6 & 62.1 & 52.6 & 44.3 & 33.4 & 22.5 & 12.7 & 42.2\\
    & \textbf{\system (Ours)} & - & \textbf{71.2} & \textbf{65.5} & \textbf{57.1} & \textbf{46.8} & \textbf{36.6} & \textbf{25.2} & \textbf{13.4} &  \textbf{45.1}\\
    \bottomrule
    \end{tabular*}
}
\label{tab:sota_thu}
\vspace{-0.05in}
\end{table*}

\vspace{-0.05in}
\section{Experiment}
\vspace{-0.05in}
\subsection{Experimental Setup}
\vspace{-0.15in}
\paragraph{Dataset}
We evaluate our method on two popular action localization datasets: THUMOS-14~\cite{idrees2017thumos} and ActivityNet-v1.3~\cite{caba2015activitynet}.
\textbf{THUMOS-14} contains untrimmed videos from 20 categories.
The video length varies from a few seconds to several minutes and multiple action instances may exist in a single video.
Following previous works~\cite{wang2017untrimmednets,paul2018w,zhai2020two,zhang2021cola}, we use the 200 videos in the validation set for training and the 213 videos in the testing set for evaluation.
\textbf{ActivityNet-v1.3} is a large-scale dataset with 200 complex daily activities. It has 10,024 training videos and 4,926 validation videos. 
Following~\cite{luo2021action,yang2021uncertainty}, we use the training set to train our model and the validation set for evaluation.

\vspace{-0.1in}
\paragraph{Implementation Details}
We employ the I3D~\cite{carreira2017quo} network pretrained on Kinetics-400~\cite{carreira2017quo} for feature extraction. 
We apply TVL1~\cite{duval2009tvl1} algorithm to extract optical flow from RGB frames. 
The Adam optimizer is used with the learning rate of 0.0001 and with the mini-batch sizes of 16, 64 for THUMOS-14 and ActivityNet-v1.3, respectively.
The number of sampled snippets $T$ is 750 for THUMOS-14 and 150 for ActivityNet-v1.3.
For the multi-step proposal refinement, $E$ is set to 100 and 50 epochs for THUMOS-14 and ActivityNet-v1.3, respectively. Action proposals are generated at the last epoch of each refinement step.
More dataset-specific training and testing details are available in the supplementary material.

\begin{algorithm}[t]
    \caption{Multi-step Proposal Refinement}
    \label{alg:refine}
    \KwIn{Training epochs $E$, refinement steps $L$}
    \KwOut{Action proposals $\tilde{\mathcal{S}}$}
    \SetAlgoLined
    Train the base model for $E$ epochs. \\
    Get initial action proposals: $\tilde{\mathcal{S}}_0$. \\
    \For{l in $\{1,...,L\}$}
    {
        Train \system for $E$ epochs with $\tilde{\mathcal{S}}_{l-1}$. \\
        Update action proposals with $\tilde{\mathcal{S}}_l$. \\
    }
\end{algorithm}%

\vspace{-0.02in}
\subsection{Comparison with the State of the Art}
\vspace{-0.05in}
In Table~\ref{tab:sota_thu}, we compare our \system with state-of-the-art 
WTAL methods on THUMOS-14. Selected fully-supervised methods are presented for reference.
We observe that \system outperforms all the previous WTAL methods and establishes new state of the art on THUMOS-14 with 45.1\% average mAP for IoU thresholds 0.1:0.7.
In particular, our approach outperforms UGCT~\cite{yang2021uncertainty}, which also utilizes pseudo labels to guide the model training but without explicit segment modeling.
Even compared with the fully supervised methods, \system outperforms SSN~\cite{zhao2017temporal} and TAL-Net~\cite{chao2018rethinking} and achieves comparable results with GTAN~\cite{long2019gaussian} and P-GCN~\cite{zeng2019graph} when the IoU threshold is low.
The results demonstrate the superior performance of our approach with action-aware segment modeling.

We also conduct experiments on ActivityNet-v1.3 and the comparison results are summarized in Table~\ref{tab:sota_anet}. Again, our \system obtains a new state-of-the-art performance of 25.1\% average mAP, surpassing the latest works (\eg UGCT~\cite{yang2021uncertainty}, FAC-Net~\cite{huang2021foreground}).
The consistent superior results on both datasets justify the effectiveness of our \system.

\begin{table}[t]
\centering
\caption{Comparison with state-of-the-art methods on ActivityNet-v1.3 dataset. The AVG column shows the averaged mAP under the IoU thresholds [0.5:0.05:0.95].}
\vspace{-0.05in}
\resizebox{0.99\linewidth}{!}{
    \renewcommand{\arraystretch}{1.2}
    \begin{tabular}{@{}llccccc@{}}
        \toprule
        \multirow{2}{*}{\textbf{Method}} & \multicolumn{1}{c}{\multirow{2}{*}{\textbf{Publication}}} & \multicolumn{4}{c}{\textbf{mAP@IoU (\%)}}\\
        \cmidrule(l{8pt}){3-6}
        & & 0.5 & 0.75 & 0.95 & AVG \\
        \midrule
        STPN~\cite{nguyen2018weakly}        & \emph{CVPR 2018} & 29.3 & 16.9 & 2.6 & 16.3 \\
        ASSG ~\cite{zhang2019adversarial}   & \emph{MM 2019}   & 32.3 & 20.1 & 4.0 & 18.8 \\
        CMCS~\cite{liu2019completeness}     & \emph{CVPR 2019} & 34.0 & 20.9 & 5.7 & 21.2 \\
        Bas-Net~\cite{lee2020background}    & \emph{AAAI 2020} & 34.5 & 22.5 & 4.9 & 22.2 \\
        TSCN~\cite{zhai2020two}             & \emph{ECCV 2020} & 35.3 & 21.4 & 5.3 & 21.7 \\
        A2CL-PT~\cite{min2020adversarial}   & \emph{ECCV 2020} & 36.8 & 22.0 & 5.2 & 22.5 \\
        ACM-Net~\cite{qu2021acm}             & \emph{TIP 2021}  & 37.6 & 24.7 & \bf 6.5 & 24.4 \\
        TS-PCA~\cite{yang2021uncertainty}   & \emph{CVPR 2021} & 37.4 & 23.5 & 5.9 & 23.7 \\
        UGCT~\cite{yang2021uncertainty}     & \emph{CVPR 2021} & 39.1 & 22.4 & 5.8 & 23.8 \\
        AUMN~\cite{luo2021action}           & \emph{CVPR 2021} & 38.3 & 23.5 & 5.2 & 23.5 \\
        FAC-Net~\cite{huang2021foreground}  & \emph{ICCV 2021} & 37.6 & 24.2 & 6.0 & 24.0 \\
        \hline
        \textbf{\system(ours)}              & &\bf 41.0 &\bf 24.9 & 6.2 &\bf 25.1 \\
        \bottomrule
    \end{tabular}
}
\label{tab:sota_anet}
\vspace{-0.2in}
\end{table}

\subsection{Ablation Studies on THUMOS-14}
\label{sec:ablation}

\begin{table*}[t]
\centering
\vspace{-0.3in}
\begin{minipage}{.38\textwidth}
    \caption{Contribution of each component. $\mathcal{L}_\text{fg}$, $\mathcal{L}_\text{bg}$ and $\mathcal{L}_\text{abg}$ represents the foreground, background and action-aware background loss, which are based on MIL with video-level labels. While DSS, Intra, Inter, and $\mathcal{L}_\text{ins}$ denote the dynamic segment sampling, intra-segment attention, inter-segment attention, and pseudo instance-level loss, respectively, which exploit segment-level information.}
    \resizebox{\linewidth}{!}{
    \renewcommand{\arraystretch}{1.2}
    \begin{tabular}{@{}ccc|cccc|c@{}}
        \toprule
        \multicolumn{3}{c|}{\textbf{Base model}} &
        \multicolumn{4}{c|}{\textbf{ASM-Loc}} &
        \textbf{AVG} \\
        \cmidrule{1-3} \cmidrule{4-7} \cmidrule{8-8} 
        $\mathcal{L}_\text{fg}$ & $\mathcal{L}_\text{bg}$ & $\mathcal{L}_\text{abg}$ & DSS & Intra & Inter & $\mathcal{L}_\text{ins}$ & 0.1:0.7 \\
        \midrule
        \checkmark & & & & & & & 24.3\\
        \checkmark & \checkmark & & & & & & 36.6 \\
        \checkmark & \checkmark & \checkmark & & & & & 40.3 \\
        \midrule
        \checkmark & \checkmark & \checkmark & \checkmark & & & & 41.4 \\
        \checkmark & \checkmark & \checkmark & & \checkmark & & & 41.8 \\
        \checkmark & \checkmark & \checkmark & & & \checkmark & & 42 \\
        \checkmark & \checkmark & \checkmark & & & & \checkmark & 41.3 \\
        \checkmark & \checkmark & \checkmark & & \checkmark & \checkmark & & 42.7  \\
        \checkmark & \checkmark & \checkmark & \checkmark & \checkmark & \checkmark & & 43.7  \\
        \checkmark & \checkmark & \checkmark &  & \checkmark & \checkmark & \checkmark & 44.3  \\
        \checkmark & \checkmark & \checkmark & \checkmark & \checkmark & \checkmark & \checkmark & \bf 45.1\\
        \bottomrule
    \end{tabular}
    }
    \label{tab:components}
\end{minipage}
\hfill
\begin{minipage}{.36\textwidth}
    \caption{Ablation on self-attention under different settings. ``Global", ``BG" indicate self-attention on all and background snippets, respectively.}
    \resizebox{\linewidth}{!}{
        \begin{tabular}{@{}cccccccc@{}}
        \toprule
        \multirow{2}{*}{\textbf{Label}} & \multirow{2}{*}{\textbf{Setting}} & \multicolumn{5}{c}{\textbf{mAP@IoU (\%)}} \\
        \cmidrule(lr){3-7}
        & & 0.1 & 0.3 & 0.5 & 0.7 & AVG \\
        \midrule
        & Base & 67.8 & 51.8 & 30.7 & 10.1 & 40.3\\
        & Global & 67.3 & 50.8 & 30.2 & 10.5 & 40.1\\
        \midrule
        \multirow{2}{*}{\tabincell{c}{Action\\Proposal}} & BG & 66 & 50.1 & 30.6 & 10.4 & 39.6 \\
        & Ours & 68.6 & 53.4 & 32.5 & 11.8 & \bf 41.8 \\
        \midrule
        \multirow{2}{*}{\tabincell{c}{Ground\\Truth}} & BG & 64.7 & 49.6 & 30.3 & 9.7 & 38.8 \\
        & Ours & 73.3 & 56.2 & 33.6 & 13.2 & 44.3 \\
        \bottomrule
        \end{tabular}
    }
    \label{tab:intra}

    \caption{Impact of dynamic segment sampling (DSS). Actions are divided into five duration groups (seconds): XS (0, 1], S (1, 2], M (2, 4], L (4, 6], and XL (6, inf).}
    \resizebox{\linewidth}{!}{
        \begin{tabular}{@{}cccccccc@{}}
        \toprule
        \multirow{2}{*}{\textbf{Label}} & \multirow{2}{*}{\textbf{Setting}} & \multicolumn{6}{c}{\textbf{Averaged mAP (\%)}} \\
        \cmidrule(lr){3-8}
        & & XS & S & M & L & XL & AVG \\
        \midrule
        & Base & 10.6 & 33.7 & 45.9 & 48.3 & 38.3 & 40.3 \\
        \midrule
        \multirow{2}{*}{\tabincell{c}{Action\\Proposal}} & +DSS & 15.5 & 34.9 & 47.1 & 48.6 & 38.5 & \bf 41.4 \\
        & $\bigtriangleup$ & \textbf{+4.9} & +1.2 & +1.2 & +0.3 & +0.2 & +1.1\\
        \midrule
        \multirow{2}{*}{\tabincell{c}{Ground\\Truth}} & +DSS & 20 & 38 & 47.6 & 49.7 & 38.8 & 43\\
        & $\bigtriangleup$ & +9.4 & +4.3 & +1.7 & +1.4 & +0.5 & +2.7\\
        \bottomrule
        \end{tabular}
    }
    \label{tab:dynamic_sampling}
\end{minipage}
\hfill
\begin{minipage}{.22\textwidth}
    \caption{Effectiveness of the uncertainty estimation module.}
    \resizebox{\linewidth}{!}{
    \renewcommand{\arraystretch}{1.2}
        \begin{tabular}{@{}ccccc@{}}
        \toprule
        \multirow{2}{*}{\textbf{Uncer.}} & \multicolumn{4}{c}{\textbf{mAP@IoU (\%)}} \\
        \cmidrule(lr){2-5}
        & 0.3 & 0.5 & 0.7 & AVG \\
        \midrule
         & 55.5 & 35.5 & 13.8 & 44.1 \\
        \checkmark & 57.1 & 36.6 & 13.4 & \bf 45.1\\
        \bottomrule
        \end{tabular}
    }
    \label{tab:uncertainty}

    \vspace{0.2in}
    \caption{Ablation on the number of refinement steps. ``0" indicates the base model without action-aware segment modeling.}
    \resizebox{\linewidth}{!}{
        \renewcommand{\arraystretch}{1.2}
        \begin{tabular}{@{}ccccc@{}}
        \toprule
        \multirow{2}{*}{\textbf{Num.}} & \multicolumn{4}{c}{\textbf{mAP@IoU (\%)}} \\
        \cmidrule(lr){2-5}
        & 0.3 & 0.5 & 0.7 & AVG \\
        \midrule
        0 & 51.8 & 30.7 & 10.1 & 40.3 \\ \midrule
        1 & 54.4 & 34.1 & 12.5 & 43.1 \\
        2 & 56.2 & 35.4 & 13.8 & 44.7 \\
        3 & 57.1 & 36.6 & 13.4 & \bf 45.1 \\
        4 & 57.3 & 36.7 & 14.1 & \bf 45.1\\
        \bottomrule
        \end{tabular}
    }
    \label{tab:iterative}
\end{minipage}
\end{table*}

\vspace{-0.15in}
\paragraph{Contribution of each component}
In Table~\ref{tab:components}, we conduct an ablation study to investigate the contribution of each component in \system.
We first observe that adding the background loss $\mathcal{L}_\text{bg}$ and the action-aware background loss $\mathcal{L}_\text{abg}$ largely enhance the performance of the base model. 
The two losses encourage the sparsity in the foreground attention weights by pushing the background attention weights to be 1 at background snippets, and therefore improve the foreground-background separation.

For action-aware segment modeling, it is obvious that a consistent gain ($\geq$1\%) can be achieved by adding any of our proposed modules. 
In particular, introducing segment modeling in the feature modeling stage (i.e., intra- and inter-segment attention) significantly increases the performance by 2.4\%. The two attention modules are complementary to each other, focusing on modeling temporal structure within and across action segments.
When incorporating all the action-aware segment modeling modules together, our approach boosts the final performance from 40.3\% to 45.1\%.

\vspace{-0.1in}
\paragraph{Are action proposals necessary for self-attention?}
We propose an intra-segment attention module that performs self-attention \textit{within} action proposals to suppress the noise from background snippets. 
To verify the effectiveness of our design, we compare different settings for self-attention in Table~\ref{tab:intra}. Specifically, the ``Global" setting indicates that the self-attention operation is applied directly to all snippets in the untrimmed video. It can be observed that this setting does not provide any gain to the baseline, as the model fails to capture meaningful temporal structure due to the existence of irrelevant and noisy background snippets.
Moreover, the ``BG" setting, which stands for self-attention on background snippets only, has negative impact and achieves even worse localization results.
Finally, our intra-segment attention outperforms these two settings by a large margin, indicating the importance of applying self-attention within action proposals.
We also present the settings of using the ground-truth action segments as proposals for intra-segment attention.
This setting can be viewed as an upper bound of our approach and it provides even more significant gains over the baseline. This observation inspires us to further improve the action proposals by multi-step refinement.

\vspace{-0.1in}
\paragraph{Impact of dynamic segment sampling}
In Table~\ref{tab:dynamic_sampling}, we evaluate the impact of dynamic segment sampling for action segments with different durations. 
We divide all action segments into five groups according to their duration in seconds and evaluate the averaged mAP~\cite{lin2014microsoft} separately for each group.
As mentioned in the introduction, localization performance on short actions (XS, S) is much worse than longer actions (M, L, XL).
By up-sampling the short actions with our dynamic segment sampling module, the model achieves significant gains on short actions (+4.9\% for XS and +1.2\% for S) and improves the overall performance by 1.1\%.
Similarly, we present the results using ground-truth segment annotation for dynamics segment sampling, which achieves even larger improvement over the baseline.

\vspace{-0.1in}
\paragraph{Impact of uncertainty estimation}
We propose an uncertainty estimation module to mitigate the noisy label problem in pseudo instance-level supervision. Table~\ref{tab:uncertainty} shows that using uncertainty estimation consistently improves the localization performance at different IoU thresholds, and increases the average mAP by 1\%.

\vspace{-0.1in}
\paragraph{Impact of multi-step refinement}
Table~\ref{tab:iterative} shows the results of increasing the number of refinement steps for multi-step proposal refinement.
We can see that the performance improves as the number of steps increases, indicating that better localization results can be achieved by refined proposals.
We adopt 3 refinement steps as our default setting since the performance saturates after that.

\subsection{Qualitative Results}
\begin{figure}[t!]
    \centering
    \adjincludegraphics[width=\linewidth, trim={{0.03\width} {0.1\height} {0.02\width} {0.1\height}},clip]{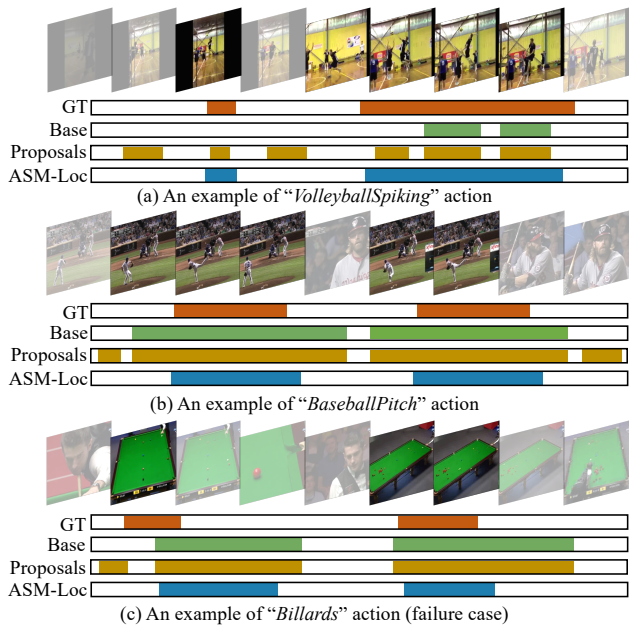}
    \vspace{-0.4in}
    \caption{Visualization of ground-truth, predictions and action proposals. Top-2 predictions with the highest confidence scores are selected for the base model and our \system. Transparent frames represent background frames.}
    \label{fig:qualitative}
    \vspace{-0.1in}
\end{figure}

Figure~\ref{fig:qualitative} shows the visualization comparisons between the base model and our \system. 
We observe that the common errors in existing MIL-based methods can be partly addressed by our action-aware segment modeling method, such as the missed detection of short actions and incomplete localization of the action ``\textit{VolleySpiking}" (Figure~\ref{fig:qualitative}(a)) and the over-complete localization of the action ``\textit{BaseballPitch}" (Figure~\ref{fig:qualitative}(b)).
We also provide a failure case in Figure~\ref{fig:qualitative}(c), where our method fails to localize the first action segment due to the largely misaligned action proposal generated by the base model. 
This also verifies the importance of improving the quality of action proposals and should be further studied in future work.

\section{Conclusion}
In this paper, we propose a novel WTAL framework named \system which enables explicit action-aware segment modeling beyond previous MIL-based methods.
We introduce three novel segment-centric modules corresponding to the three stages of a WTAL pipeline, which narrows the performance gap between the weakly-supervised and fully-supervised settings.
We further introduce a multi-step training strategy to progressively refine the action proposals till the localization performance saturates.
Our \system achieves state-of-the-art results on two WTAL benchmarks.

\medskip
\noindent\textbf{Acknowledgements.} This work was supported by the Air Force (STTR awards FA865019P6014, FA864920C0010) and Amazon Research Award to AS.
\clearpage
{
    \small
    \bibliography{macros,main}
}

\clearpage
\appendix
\section*{Appendix}

Sec.~\ref{sec:extra} reports additional experiments and analysis.
Sec.~\ref{sec:proposal} elaborates on the procedure of action proposal generation.
Sec.~\ref{sec:experiment_details} provides more dataset-specific implementation details and hyper-parameters for training and testing.
We also provide more qualitative results in Sec.~\ref{sec:qualitative_more}.
We discuss the limitation and broader impact of our work in Sec.~\ref{sec:limitation} and Sec.~\ref{sec:impact}.

\section{Additional Experiments and Analysis}
\label{sec:extra}

\begin{figure}[t]
    \centering
    \vspace{-0.1in}
    \adjincludegraphics[width=\linewidth, trim={{0.05\width} {0.1\height} {0.05\width} {0.1\height}},clip]{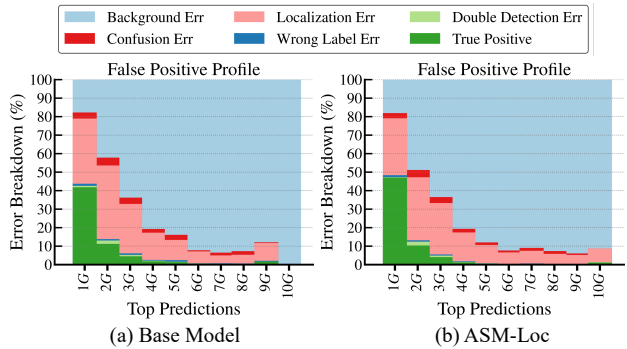}
    \vspace{-0.4in}
    \caption{Diagnosing detection results. We present DETAD~\cite{alwassel_2018_detad} false positive profiles of the base model and our \system.}
    \label{fig:error}
\end{figure}

\paragraph{Error analysis}
To analyze the effectiveness of our \system, we conduct a DETAD~\cite{alwassel_2018_detad} false positive analysis of the base model without any action-aware segment modeling modules and our \system. We present the results in Figure~\ref{fig:error}. It shows a detailed categorization of false positive errors and summarizes the distribution of these errors.
$G$ represents the number of ground truth segments in the THUMOS-14 dataset. 
We can observe that \system generates more true positive predictions with high confidence scores and produces less localization error and confusion error (at the top 1$G$ scoring predictions).
It verifies that \system improves the detection results by predicting more accurate action boundaries with our action-aware segment modeling modules.

\paragraph{Ablation on the increased receptive field}
To further demonstrate that the effectiveness of our intra- and inter-segment attention modules is due to the segment-centric design instead of the increased receptive field, we replace our intra- and inter-segment attention modules with convolutional layers and compare the experimental results. From Table~\ref{tab:receptive} we can see that by replacing the attention modules with convolutional layers, the performances drop by 
at least $3.3\%$, and even fall below the base model. We hypothesize that increasing the kernel size of the convolutional layers may lead to confusion between foreground and background snippets especially near the action boundaries.
In contrast, our segment-centric attention design can model temporal structures within and across action segments and localize actions more precisely.
The results verify that the segment-centric design is the key to our intra- and inter-segment attention modules.

\begin{algorithm*}[t!]
    \caption{Action Proposal Generation}
    \label{alg:generation}
    \KwIn{Predicted Action Segments $\mathcal{S} = \{(s_i, e_i, c_i, q_i)\}_{i=1}^{I}$, selection ratio $\alpha$, segment extension parameter $\delta$}
    \KwOut{Action Proposals $\tilde{\mathcal{S}} = \{(\tilde{s}_n, \tilde{e}_n, \tilde{c}_n)\}_{n=1}^N$}
    \SetAlgoLined
    \For{ground-truth class $c$}
    {
        $\mathcal{S}(c)_{sorted} \leftarrow$ SORT$(\mathcal{S}(c))$ \COMMENT{sort segments by scores of class $c$}\\
        $q_{sum} = \sum q_i$ \COMMENT{sum confidence scores for all segments} \\
        Select $K$, s.t. $\max_{K} \sum_{i=1}^{K} q_i \leq \alpha * q_{sum}$ \COMMENT{select top-$K$ segments from $\mathcal{S}(c)_{sorted}$}\\
        $\tilde{\mathcal{S}}(c): \{\tilde{s}_i, \tilde{e}_i, \tilde{c}_i\}_{i=1}^K = \{s_i - \delta(e_i - s_i), e_i + \delta(e_i - s_i), c_i\}_{i=1}^K$ \COMMENT{extend selected segments on both sides}\\
    }
\end{algorithm*}

\begin{table}[t]
\centering
\caption{Ablation on the increased receptive field.}
\resizebox{0.99\linewidth}{!}{
    \begin{tabular}{@{}cccccccccc@{}}
    \toprule
    \multirow{2}{*}{\textbf{Modeling}} & \multirow{2}{*}{\textbf{\tabincell{c}{Kernel\\Size}}} & \multicolumn{8}{c}{\textbf{mAP@IoU (\%)}} \\
    \cmidrule(l{4pt}r{1pt}){3-10}
    & & 0.1 & 0.2 & 0.3 & 0.4 & 0.5 & 0.6 & 0.7 & AVG \\
    \midrule
    Base & - & 67.8 & 60.7 & 51.8 & 41.3 & 30.7 & 19.9 & 10.1 & 40.3 \\
    \midrule
     & 3 & 66.2 & 59.3 & 50.5 & 39.9 & 29.9 & 19.2 & 9.1 & 39.2 \\
    \multirow{2}{*}{Conv} & 5 & 66.5 & 58.9 & 51.0 & 40.0 & 29.7 & 19.3 & 9.8 & 39.3 \\
     & 9 & 67.1 & 59.8 & 50.4 & 40.1 & 29.1 & 19.2 & 10.2 & 39.4 \\
     \midrule
    Attention & - & 68.9 & 63.1 & 54.9 & 44.5 & 34 & 22.0 & 11.9 & \bf 42.7 \\
    \bottomrule
    \end{tabular}
}
\label{tab:receptive}
\end{table}

\section{Action Proposal Generation}
\label{sec:proposal}
\begin{table}[t]
\centering
\caption{Ablation on different action proposal selection methods.}
\resizebox{0.99\linewidth}{!}{
    \begin{tabular}{@{}ccccccccc@{}}
    \toprule
    \multirow{2}{*}{\textbf{Method}} & \multicolumn{8}{c}{\textbf{mAP@IoU (\%)}} \\
    \cmidrule(l{4pt}r{1pt}){2-9}
    & 0.1 & 0.2 & 0.3 & 0.4 & 0.5 & 0.6 & 0.7 & AVG \\
    \midrule
    (a) & 69.9 & 63.8 & 56 & 45.8 & 36.6 & 25.0 & 13.5 & 44.4 \\
    (b) & 70.5 & 64.6 & 57.3 & 46.8 & 35.7 & 24.3 & 14.2 & 44.8 \\
    (c) & 71.2 & 65.5 & 57.1 & 46.8 & 36.6 & 25.2 & 13.4 & \bf 45.1 \\
    \bottomrule
    \end{tabular}
}
\vspace{-0.1in}
\label{tab:selection}
\end{table}

In Alg.~\ref{alg:generation}, we present the details of how to generate action proposals $\tilde{\mathcal{S}}$ from the action localization results (\ie, action segments) $\mathcal{S}$.
Specifically, we first sort all the segment scores across the set $\mathcal{S}(c)$ for each ground-truth class $c$. 
Then we sum the confidence scores of all the action segments and output $q_{sum}$, and pick the top-$K$ action segments with their confidence scores summation equal to $\alpha * q_{sum}$ to form the action proposals.
Note that the number of the action proposals is video-adaptive and content dependent, despite $\alpha$ is shared for all videos.
Finally, following the common practice in temporal action localization~\cite{shou2017cdc,lin2018bsn,zeng2019graph,lin2020fast}, we extend each proposal on both ends by $\delta$ of the proposal length to get an extended proposal with a longer temporal duration which can take more context-related snippets into consideration.

To verify the effectiveness of our proposal generation design, we compare three different settings of the segment selection procedure:
\textbf{(a)} Fixed number of selected action segments where $K$ is a fixed value for each class, which is not video-adaptive and content dependent; 
\textbf{(b)} $K$ proportional to the number of predicted action segments in $\mathcal{S}(c)$, where $K = \alpha * |\mathcal{S}(c)|$;
\textbf{(c)} our design.
In Table~\ref{tab:selection}, we can see that our design achieves the best results among the three designs.

\section{Experiment Details}
\label{sec:experiment_details}
For the hyper-parameters, we set $\lambda_\text{fg}=1, \lambda_\text{bg}=0.5, \lambda_\text{abg}=0.5, \beta=0.2, \gamma=6, H=8, \delta=0.5, \alpha=0.7$ for THUMOS-14 and $\lambda_\text{fg}=5, \lambda_\text{bg}=0.5, \lambda_\text{abg}=0.5, \beta=0.2, \gamma=10, H=8, \delta=0, \alpha=0.3$ for ActivityNet-v1.3.

Following~\cite{zhang2021cola,qu2021acm}, during inference, we use a set of thresholds to obtain the predicted action instances, then perform non-maximum suppression to remove overlapping segments. Specifically, for THUMOS-14, we set the foreground attention threshold from 0.1 to 0.9 with step 0.025, and perform NMS with a t-IOU threshold of 0.45. For ActivityNet-v1.3, we set the foreground-attention threshold from 0.005 to 0.02 with step 0.005, and apply NMS with a t-IoU threshold of 0.9.

We implement our method in PyTorch~\cite{paszke2019pytorch} and train it on a single NVIDIA RTX1080Ti gpu.

\begin{figure}[t]
    \centering
    \vspace{-0.2in}
    \adjincludegraphics[width=\linewidth, trim={{0.03\width} {0.1\height} {0.02\width} {0.1\height}},clip]{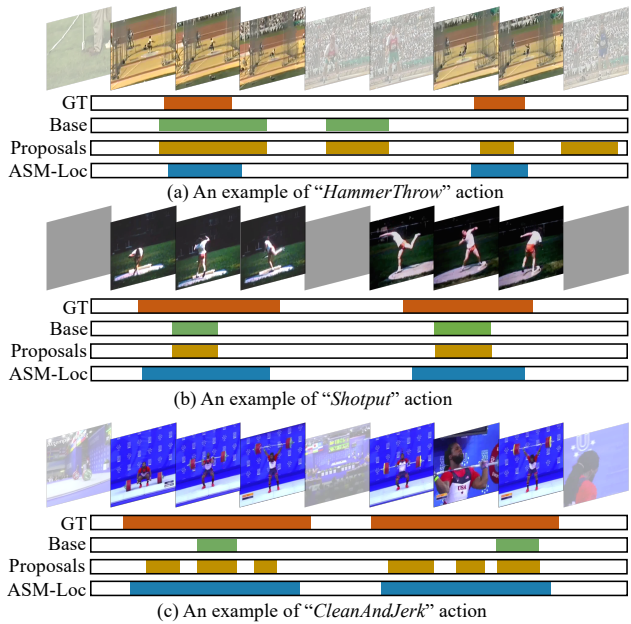}
    \vspace{-0.2in}
    \caption{Visualization of ground-truth, predictions and action proposals. Top-2 predictions with the highest confidence scores are selected for the base model and our \system. Transparent frames represent background frames.}
    \label{fig:qualitative_more}
    \vspace{-0.1in}
\end{figure}

\section{More Qualitative Results}
\label{sec:qualitative_more}
We provide more qualitative results in Figure~\ref{fig:qualitative_more}. 
The first example of action ``\textit{HammerThrow}" shows the missed detection of short actions and over-completeness error.
The second and third example of action ``\textit{Shotput}" and action ``\textit{CleanAndJerk}" shows the incompleteness error.
It clearly shows that our \system can help address these errors with more accurate action boundary predictions.

\section{Limitation}
\label{sec:limitation}
The main limitation of our \system is that the performance of our action-aware segment modeling modules depends on the generated action proposals. When the action proposals are largely misaligned with the ground-truth action segments, our \system is not able to fix the error and generate correct predictions, as shown in Figure~\textcolor{red}{3}.

\section{Broader Impacts}
\label{sec:impact}
As the most popular media format nowadays, most information is spread in the format of videos. 
The temporal action localization task aims at finding the temporal boundaries and classifying category labels of actions of interest in untrimmed videos. 
Unlike supervised learning based approach that requires dense segment-level annotations, our proposed weakly-supervised temporal action localization model \system only requires video-level labels. 
Therefore, WTAL is much more valuable in the real-world applications such as popular video-sharing social-network services, where billions of videos have only video-level user-generated tags.
Besides, WTAL has broad applications in various fields, \eg event detection, video summarization, highlight generation and video surveillance.

\end{document}